\documentclass[]{fairmeta}
\usepackage{wrapfig}
\usepackage{hyperref}
\usepackage{url}
\usepackage{algcompatible}
\usepackage{algorithm}
\usepackage{algpseudocode}
\usepackage{graphicx}
\usepackage[T1]{fontenc} 
\usepackage{booktabs}
\usepackage{multirow}
\usepackage{xspace}
\usepackage{float}
\usepackage{fontawesome5}
\newcommand{\ours}{\textsc{Coconut}\xspace}
\NewDocumentCommand{\shibo}{ mO{} }{\textcolor{cyan}
{\textsuperscript{\textit{Shibo}}\textsf{\textbf{\small[#1]}}}}

\NewDocumentCommand{\andy}{ mO{} }{\textcolor{purple}
{\textsuperscript{\textit{Andy}}\textsf{\textbf{\small[#1]}}}}
\newcommand{\dataset}{ProsQA}

\usepackage{amsmath,amsfonts,bm}

\def\eqref#1{equation~\ref{#1}}

\def\1{\bm{1}}

\DeclareMathAlphabet{\mathsfit}{\encodingdefault}{\sfdefault}{m}{sl}
\SetMathAlphabet{\mathsfit}{bold}{\encodingdefault}{\sfdefault}{bx}{n}

\newcommand{\softmax}{\mathrm{softmax}}

\title{Training Large Language Models to Reason in a Continuous Latent Space}

\author[1,2,*]{Shibo Hao}
\author[1]{Sainbayar Sukhbaatar}
\author[1]{DiJia Su}
\author[1]{Xian Li}
\author[2]{Zhiting Hu}
\author[1]{Jason Weston}
\author[1]{Yuandong Tian}

\affiliation[1]{FAIR at Meta}
\affiliation[2]{UC San Diego}

\contribution[*]{Work done at Meta}

\abstract{
Large language models (LLMs) are restricted to reason in the ``language space'', where they typically express the reasoning process with a chain-of-thought (CoT) to solve a complex reasoning problem. However, we argue that language space may not always be optimal for reasoning. For example, most word tokens primarily ensure textual coherence and are not essential for reasoning, while some critical tokens require complex planning and pose huge challenges to LLMs. To explore the potential of LLM reasoning in an unrestricted latent space instead of using natural language, we introduce a new paradigm \ours (\underline{C}hain \underline{o}f \underline{Con}tin\underline{u}ous \underline{T}hought). We utilize the last hidden state of the LLM as a representation of the reasoning state (termed ``continuous thought''). Rather than decoding this into a word token, we feed it back to the LLM as the subsequent input embedding directly in the continuous space. This latent reasoning paradigm leads to the emergence of an advanced reasoning pattern: the continuous thought can encode multiple alternative next reasoning steps, allowing the model to perform a breadth-first search (BFS) to solve the problem, rather than prematurely committing to a single deterministic path like CoT. \ours outperforms CoT on certain logical reasoning tasks that require substantial search during planning, and shows a better trade-off between accuracy and efficiency.
}

\date{\today\\
\textbf{Code}: \url{https://github.com/facebookresearch/coconut}}

\begin{document}

\maketitle

\section{Introduction}
\label{section:intro}
Large language models (LLMs) have demonstrated remarkable reasoning abilities, emerging from extensive pretraining on human languages~\citep{dubey2024llama, achiam2023gpt}. While next token prediction is an effective training objective, it imposes a fundamental constraint on the LLM as a reasoning machine: the explicit reasoning process of LLMs must be generated in word tokens. For example, a prevalent approach, known as chain-of-thought (CoT) reasoning~\citep{wei2022chain}, involves prompting or training LLMs to generate solutions step-by-step using natural language. However, this is in stark contrast to certain human cognition results. Neuroimaging studies have consistently shown that the language network -- a set of brain regions responsible for language comprehension and production -- remains largely inactive during various reasoning tasks ~\citep{amalric2019distinct, monti2012thought, monti2007functional, monti2009boundaries, fedorenko2011functional}. Further evidence indicates that human language is optimized for communication rather than reasoning~\citep{fedorenko2024language}.

A significant issue arises when LLMs use language for reasoning: the amount of reasoning required for each particular token varies greatly, yet current LLM architectures allocate nearly the same computing budget for predicting every token. Most tokens in a reasoning chain are generated solely for fluency, contributing little to the actual reasoning process. By contrast, some critical tokens require complex planning and pose huge challenges to LLMs. While previous work has attempted to fix these problems by prompting LLMs to generate succinct reasoning chains~\citep{madaan2022text}, or performing additional reasoning before generating some critical tokens~\citep{zelikman2024quiet}, these solutions remain constrained within the language space and do not solve the fundamental problems. On the contrary, it would be ideal for LLMs to have the freedom to reason without any language constraints, and then translate their findings into language only when necessary.


\begin{figure}
    \centering
    \includegraphics[width=\linewidth]{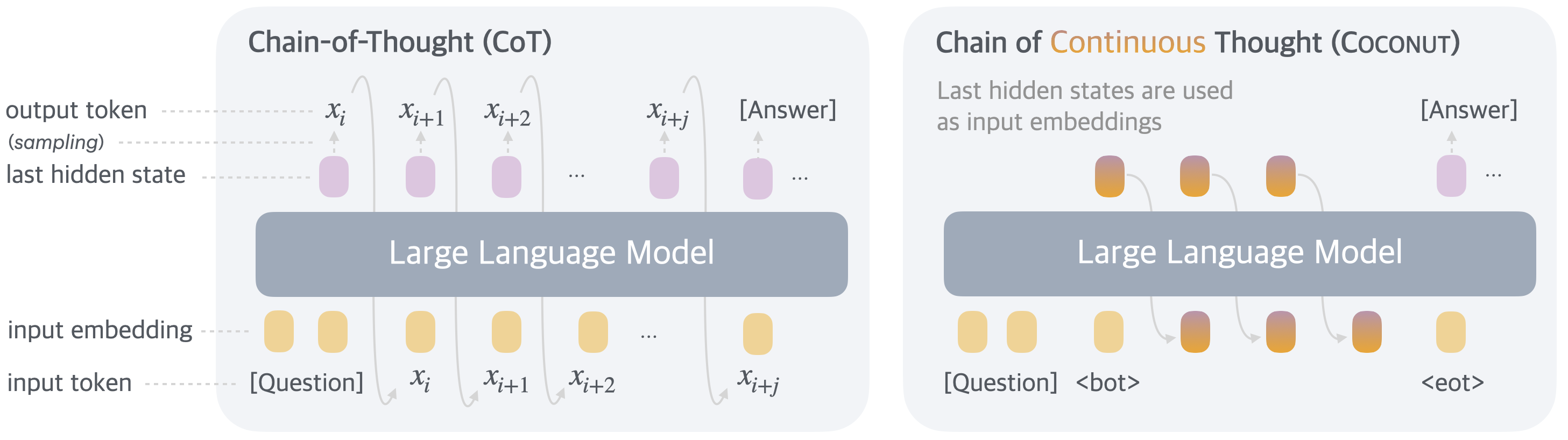}
    \caption{A comparison of Chain of Continuous Thought (\ours) with Chain-of-Thought (CoT). In CoT, the model generates the reasoning process as a word token sequence (e.g., $[x_{i}, x_{i+1}, ..., x_{i+j}]$ in the figure). \ours  regards the last hidden state as a representation of the reasoning state (termed ``continuous thought''), and directly uses it as the next input embedding. This allows the LLM to reason in an unrestricted latent space instead of a language space.}
    \label{fig:motivation}
\end{figure}

In this work we instead explore LLM reasoning in a latent space by introducing a novel paradigm, \ours (Chain of Continuous Thought). It involves a simple modification to the traditional CoT process: instead of mapping between hidden states and language tokens using the language model head and embedding layer, \ours directly feeds the last hidden state (a continuous thought) as the input embedding for the next token (Figure~\ref{fig:motivation}). This modification frees the reasoning from being within the language space, and the system can be optimized end-to-end by gradient descent, as continuous thoughts are fully differentiable. To enhance the training of latent reasoning, we employ a multi-stage training strategy inspired by \citet{deng2024explicit}, which effectively utilizes language reasoning chains to guide the training process.

Interestingly, our proposed paradigm leads to an efficient reasoning pattern. 
Unlike language-based reasoning, continuous thoughts in \ours can encode multiple potential next steps simultaneously, allowing for a reasoning process akin to breadth-first search (BFS). While the model may not initially make the correct decision, it can maintain many possible options within the continuous thoughts and progressively eliminate incorrect paths through reasoning, guided by some implicit value functions. This advanced reasoning mechanism surpasses traditional CoT, even though the model is not explicitly trained or instructed to operate in this manner, as seen in previous works \citep{yao2023tree, hao2023reasoning}.

Experimentally, \ours successfully enhances the reasoning capabilities of LLMs. For math reasoning (GSM8k,~\citealp{cobbe2021training}), using continuous thoughts is shown to be beneficial to reasoning accuracy, mirroring the effects of language reasoning chains. This indicates the potential to scale and solve increasingly challenging problems by chaining more continuous thoughts. On logical reasoning including ProntoQA~\citep{saparov2022language}, and our newly proposed ProsQA (Section~\ref{sec:understanding}) which requires stronger planning ability, \ours and some of its variants even surpasses language-based CoT methods, while generating significantly fewer tokens during inference. We believe that these findings underscore the potential of latent reasoning and could provide valuable insights for future research.

\section{Related Work}
\label{related_work}

\noindent\textbf{Chain-of-thought (CoT) reasoning.} We use the term chain-of-thought broadly to refer to methods that generate an intermediate reasoning process in language before outputting the final answer. This includes prompting LLMs~\citep{wei2022chain, khot2022decomposed, zhou2022least}, or training LLMs to generate reasoning chains, either with supervised finetuning~\citep{yue2023mammoth, yu2023metamath} or reinforcement learning~\citep{wang2024math, havrilla2024teaching, shao2024deepseekmath, yu2024flow}. 
\citet{madaan2022text} classified the tokens in CoT into symbols, patterns, and text, and proposed to guide the LLM to generate concise CoT based on analysis of their roles. Recent theoretical analyses have demonstrated the usefulness of CoT from the perspective of model expressivity~\citep{feng2023towards, merrill2023expresssive, li2024chain}. By employing CoT, the effective depth of the transformer increases because the generated outputs are looped back to the input~\citep{feng2023towards}. These analyses, combined with the established effectiveness of CoT, motivated our design of feeding the continuous thoughts back into the LLM as input embeddings. 
While CoT has proven effective for certain tasks, its autoregressive generation nature makes it challenging to mimic human reasoning on more complex problems~\citep{lecun2022path, hao2023reasoning}, which typically require planning and search. There are works that equip LLMs with explicit tree search algorithms~\citep{ xie2023self, yao2023tree, hao2024llm}, or train the LLM on search dynamics and trajectories~\citep{lehnert2024beyond, gandhi2024stream, su2024dualformer}. In our analysis, we find that after removing the constraint of a language space, a new reasoning pattern similar to BFS emerges, even though the model is not explicitly trained in this way.

\noindent\textbf{Latent reasoning in LLMs.} Previous works mostly define latent reasoning in LLMs as the hidden computation in transformers~\citep{yang2024large, biran2024hopping}. \citet{yang2024large} constructed a dataset of two-hop reasoning problems and discovered that it is possible to recover the intermediate variable from the hidden representations. \citet{biran2024hopping} further proposed to intervene the latent reasoning by ``back-patching'' the hidden representation. \citet{shalev2024distributional} discovered parallel latent reasoning paths in LLMs. Another line of work has discovered that, even if the model generates a CoT to reason, the model may actually utilize a different latent reasoning process. This phenomenon is known as the unfaithfulness of CoT reasoning~\citep{wang2022towards, turpin2024language}. To enhance the latent reasoning of LLMs, previous research proposed to augment it with additional tokens. \citet{goyal2023think} pretrained the model by randomly inserting a learnable \texttt{<pause>} tokens to the training corpus. This improves LLM's performance on a variety of tasks, especially when followed by supervised finetuning with \texttt{<pause>} tokens. On the other hand, \citet{pfau2024let} further explored the usage of filler tokens, e.g., ``\texttt{...}'', and concluded that they work well for highly parallelizable problems. However, \citet{pfau2024let} mentioned these methods do not extend the expressivity of the LLM like CoT; hence, they may not scale to more general and complex reasoning problems. \citet{wang2023guiding} proposed to predict a planning token as a discrete latent variable before generating the next reasoning step. Recently, it has also been found that one can ``internalize'' the CoT reasoning into latent reasoning in the transformer with knowledge distillation~\citep{deng2023implicit} or a special training curriculum which gradually shortens CoT~\citep{deng2024explicit}. \citet{yu2024distilling} also proposed to distill a model that can reason latently from data generated with complex reasoning algorithms. These training methods can be combined to our framework, and specifically, we find that breaking down the learning of continuous thoughts into multiple stages, inspired by iCoT~\citep{deng2024explicit}, is very beneficial for the training. Other work explores alternative architectures for latent reasoning, including looped transformers~\citep{giannou2023looped, fan2024looped}, diffusion models in sentence embedding space~\citep{barrault2024large}. Recurrent memory transformers similarly pass information across segments through continuous memory embeddings updated at each recurrent step rather than through language tokens~\citep{bulatov2022recurrent}, with associative memory mechanisms later introduced to scale this recurrence to longer contexts~\citep{rodkin2024associative}. Different from these works, we focus on general multi-step reasoning tasks and aim to investigate the unique properties of latent reasoning in comparison to language space. In addition to reasoning tasks,
\citet{pham2023let} also explored using continuous space for multi-agent communication. Building on \ours, \citet{zhu2025reasoning} developed a theoretical framework demonstrating that continuous CoT can be more efficient than discrete CoT on certain tasks by encoding multiple reasoning paths in superposition states. Subsequently, \citet{zhu2025emergence} analyzed the training dynamics to explain how such superposition emerges under the \ours\ training objective.

\section{\ours: Chain of Continuous Thought}
\label{sec:ours}
In this section, we introduce our new paradigm \ours (Chain of Continuous Thought) for reasoning in an unconstrained latent space. We begin by introducing the background and notation we use for language models. For an input sequence $x = (x_1, ..., x_T)$, the standard large language model $\mathcal{M}$ can be described as:

$$H_t = \text{Transformer}(E_t)$$
$$\mathcal{M}(x_{t+1}\mid x_{\leq t}) = \text{softmax}(Wh_t)$$

where $E_t = [e(x_1), e(x_2), ..., e(x_t)]$ is the sequence of token embeddings up to position $t$;
$H_t \in \mathbb{R}^{t \times d}$ is the matrix of the last hidden states for all tokens up to position $t$; $h_t$ is the last hidden state of position $t$, i.e., $h_t=H_t[t, :]$; $e(\cdot)$ is the token embedding function; $W$ is the parameter of the language model head.

\noindent\textbf{Method Overview.} In the proposed \ours method, the LLM switches between the ``language mode'' and ``latent mode'' (Figure~\ref{fig:motivation}). In language mode, the model operates as a standard language model, autoregressively generating the next token. In latent mode, it directly utilizes the last hidden state as the next input embedding. This last hidden state represents the current reasoning state, termed as a ``continuous thought''.

Special tokens \texttt{<bot>} and \texttt{<eot>} are employed to mark the beginning and end of the latent thought mode, respectively. As an example, we assume latent reasoning occurs between positions $i$ and $j$, i.e., $x_i=$ \texttt{<bot>} and $x_j=$ \texttt{<eot>}. When the model is in the latent mode ($i < t < j$), we use the last hidden state from the previous token to replace the input embedding, i.e., $E_t=[e(x_1), e(x_2), ..., e(x_i), h_i, h_{i+1}, ..., h_{t-1}]$. After the latent mode finishes ($t \ge j$), the input reverts to using the token embedding, i.e., $E_t=[e(x_1), e(x_2), ..., e(x_i), h_i, h_{i+1}, ..., h_{j-1}, e(x_j), ..., e(x_t)]$. 
It is worth noting that the last hidden states have been processed by the final normalization layer, so they are not too large in magnitude.
$\mathcal{M}(x_{t+1}\mid x_{\leq t})$ is not defined when $i<t<j$, since the latent thought is not intended to be mapped back to language space. However, $\softmax(Wh_t)$ can still be calculated for probing purposes (see Section~\ref{sec:experiment}).

\begin{figure}
    \centering
    \includegraphics[width=\linewidth]{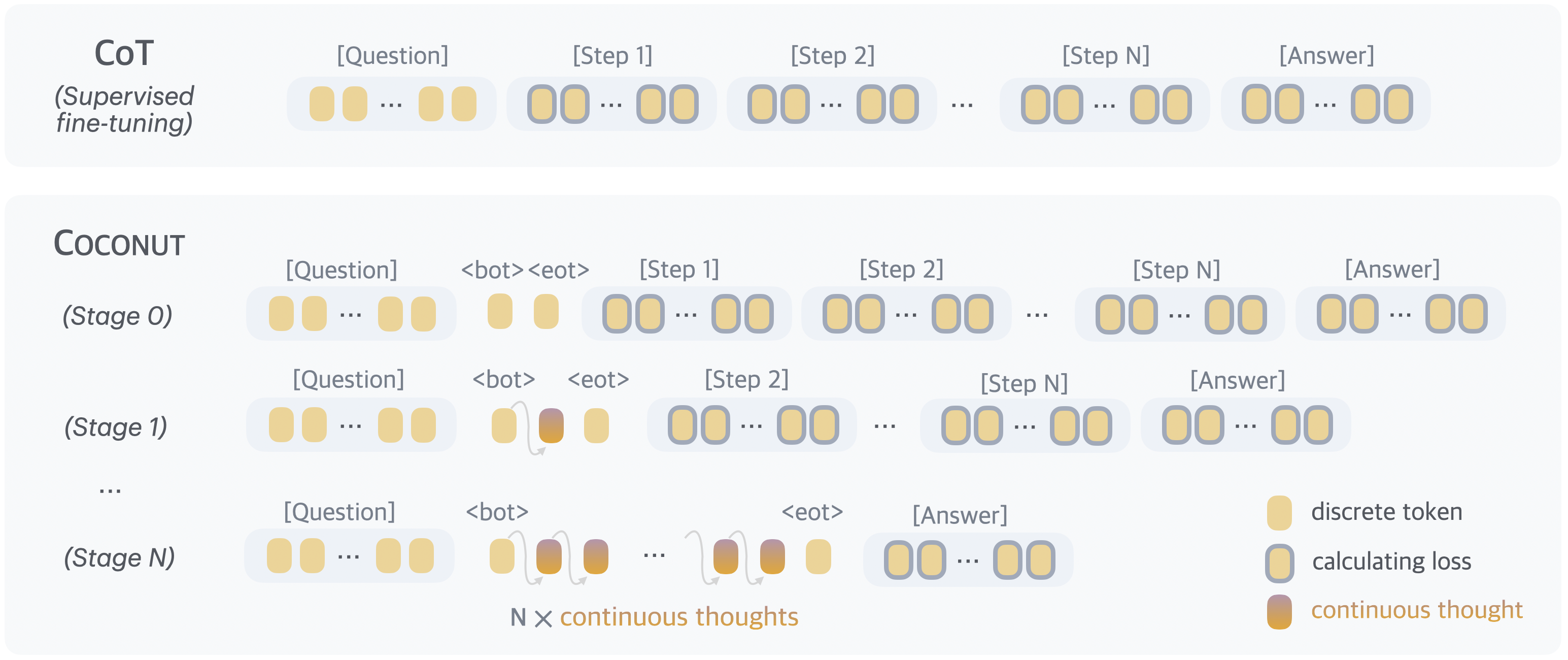}
    \caption{Training procedure of Chain of Continuous Thought (\ours). 
    Given training data with language reasoning steps, at each training stage we integrate $c$ additional continuous thoughts ($c=1$ in this example), and remove one language reasoning step. The cross-entropy loss is then used on the remaining tokens after continuous thoughts.}
    \label{fig:training}
\end{figure}

\noindent\textbf{Training Procedure.} 
In this work, we focus on a problem-solving setting where the model receives a question as input and is expected to generate an answer through a reasoning process. We leverage language CoT data to supervise continuous thought by implementing a multi-stage training curriculum inspired by \citet{deng2024explicit}. As shown in Figure~\ref{fig:training}, in the initial stage, the model is trained on regular CoT instances. In the subsequent stages, at the $k$-th stage, the first $k$ reasoning steps in the CoT are replaced with $k \times c$ continuous thoughts\footnote{If a language reasoning chain is shorter than $k$ steps, then all the language thoughts will be removed.}, where $c$ is a hyperparameter controlling the number of latent thoughts replacing a single language reasoning step. Following \citet{deng2024explicit}, we also reset the optimizer state when training stages switch. We insert \texttt{<bot>} and \texttt{<eot>} tokens (which are not counted towards $c$) to encapsulate the continuous thoughts.

During the training process, we optimize the normal negative log-likelihood loss, but mask the loss on questions and latent thoughts. It is important to note that the objective does \textbf{not} encourage the continuous thought to \textit{compress the removed language thought}, but rather to \textit{facilitate the prediction of future reasoning}. Therefore, it's possible for the LLM to learn more effective representations of reasoning steps compared to human language. 

\textbf{Training Details.} Our proposed continuous thoughts are fully differentiable and allow for back-propagation. We perform $n+1$ forward passes when $n$ latent thoughts are scheduled in the current training stage, computing a new latent thought with each pass and finally conducting an additional forward pass to obtain a loss on the remaining text sequence. While we can save any repetitive computing by using a KV cache, the sequential nature of the multiple forward passes poses challenges for parallelism. Further optimizing the training efficiency of \ours remains an important direction for future research.

\noindent\textbf{Inference Process.} The inference process for \ours is analogous to standard language model decoding, except that in latent mode, we directly feed the last hidden state as the next input embedding. A challenge lies in determining when to switch between latent and language modes. As we focus on the problem-solving setting, we insert a \texttt{<bot>} token immediately following the question tokens. For \texttt{<eot>}, we consider two potential strategies: a) train a binary classifier on latent thoughts to enable the model to autonomously decide when to terminate the latent reasoning, or b) always pad the latent thoughts to a constant length. We found that both approaches work comparably well. Therefore, we use the second option in our experiment for simplicity, unless specified otherwise.

\section{Continuous Space Enables Latent Tree Search}

\label{sec:understanding}

In this section, we provide a proof of concept of the advantage of continuous latent space reasoning. On ProsQA, a new dataset that requires extensive planning ability, \ours outperforms language space CoT reasoning. Interestingly, our analysis indicates that the continuous representation of reasoning can encode multiple alternative next reasoning steps. This allows the model to perform a breadth-first search (BFS) to solve the problem, instead of prematurely committing to a single deterministic path like language CoT.

We start by introducing the experimental setup (Section~\ref{sec:search_setup}). By leveraging \ours 's ability to switch between language and latent space reasoning, we are able to control the model to interpolate between fully latent reasoning and fully language reasoning and test their performance (Section~\ref{sec:understanding_results}). This also enables us to interpret the latent reasoning process as tree search (Section~\ref{sec:interpret}). Based on this perspective, we explain why latent reasoning can help LLMs make better decisions (Section~\ref{sec:height}).

\subsection{Experimental Setup}

\label{sec:search_setup}

\noindent\textbf{Dataset.} We introduce ProsQA (\underline{Pro}of with \underline{S}earch \underline{Q}uestion-\underline{A}nswering), a new logical reasoning dataset. A visualized example is shown in Figure~\ref{fig:interpret}. Each instance in ProsQA consists of a directed acyclic graph (DAG) of logical relationships between concepts, presented as natural language statements. The task requires models to determine logical relationships by finding valid paths through this graph, demanding sophisticated planning and search strategies. Unlike previous logical reasoning datasets like ProntoQA~\citep{saparov2022language}, ProsQA's DAG structure introduces complex exploration paths, making it particularly challenging for models to identify the correct reasoning chain. More comprehensive details about the dataset construction and characteristics can be found in Appendix~\ref{sec:dataset-appendix}.

\noindent\textbf{Setup.} We use a pre-trained GPT-2 model as the base model for all experiments. The learning rate is set to $1\times 10^{-4}$ while the effective batch size is 128. We train a \ours model following the training procedure in Section~\ref{sec:ours}. Since the maximum reasoning steps in ProsQA is 6, we set the number of training stages to $N=6$ in the training procedure. In each stage, we train the model for 5 epochs, and stay in the last stage until the 50 epochs. The checkpoint with the best accuracy in the last stage is used for evaluation. As reference, we report the performance of (1) \textit{CoT}: the model is trained with CoT data, and during inference, the model will generate a complete reasoning chain to solve the problem. (2) \textit{no-CoT}: the model is trained with only the question and answer pairs, without any reasoning steps. During inference, the model will output the final answer directly. 


To understand the properties of latent and language reasoning space, \textit{we manipulate the model to switch between fully latent reasoning and fully language reasoning}, by manually setting the position of the \texttt{<eot>} token during inference. When we enforce \ours to use $k$ continuous thoughts, the model is expected to output the remaining reasoning chain in language, starting from the $k+1$ step. In our experiments, we test variants of \ours on ProsQA with $k \in \{0, 1, 2, 3, 4, 5, 6\}$. Note that all these variants only differ in inference time while sharing the same model weights.

\noindent\textbf{Metrics.} We apply two sets of evaluation metrics. One of them is based on the correctness of the \textit{final answer}, regardless of the reasoning process. It is also the main metric used in the later sections (Section~\ref{sec:result}). To enable fine-grained analysis on ProsQA, we define another metric on the \textit{reasoning process}. We classify a reasoning chain into (1) \textbf{Correct Path}: The output is one of the shortest paths to the correct answer. (2) \textbf{Longer Path}: A valid path that correctly answers the question but is longer than the shortest path. (3) \textbf{Hallucination}: The path includes nonexistent edges or is disconnected. (4) \textbf{Wrong Target}: A valid path in the graph, but the destination node is not the one being asked. These four categories naturally apply to the output from \ours ($k=0$) and \textit{CoT}, which generate the full path. For \ours with $k>0$ that outputs only partial paths in language (with the initial steps in continuous reasoning), we classify the reasoning as a Correct Path \textit{if a valid explanation can complete it}. Also, we define Longer Path and Wrong Target for partial paths similarly. If no valid explanation completes the path, it's classified as Hallucination. In \textit{no-CoT} and \ours with larger $k$, the model may only output the final answer without any partial path, and it falls into (5) \textbf{Correct Label} or (6) \textbf{Incorrect Label}. These six categories cover all cases without overlap.

\subsection{Overall Results}

\label{sec:understanding_results}

\begin{figure}
    \centering
    \includegraphics[width=\linewidth]{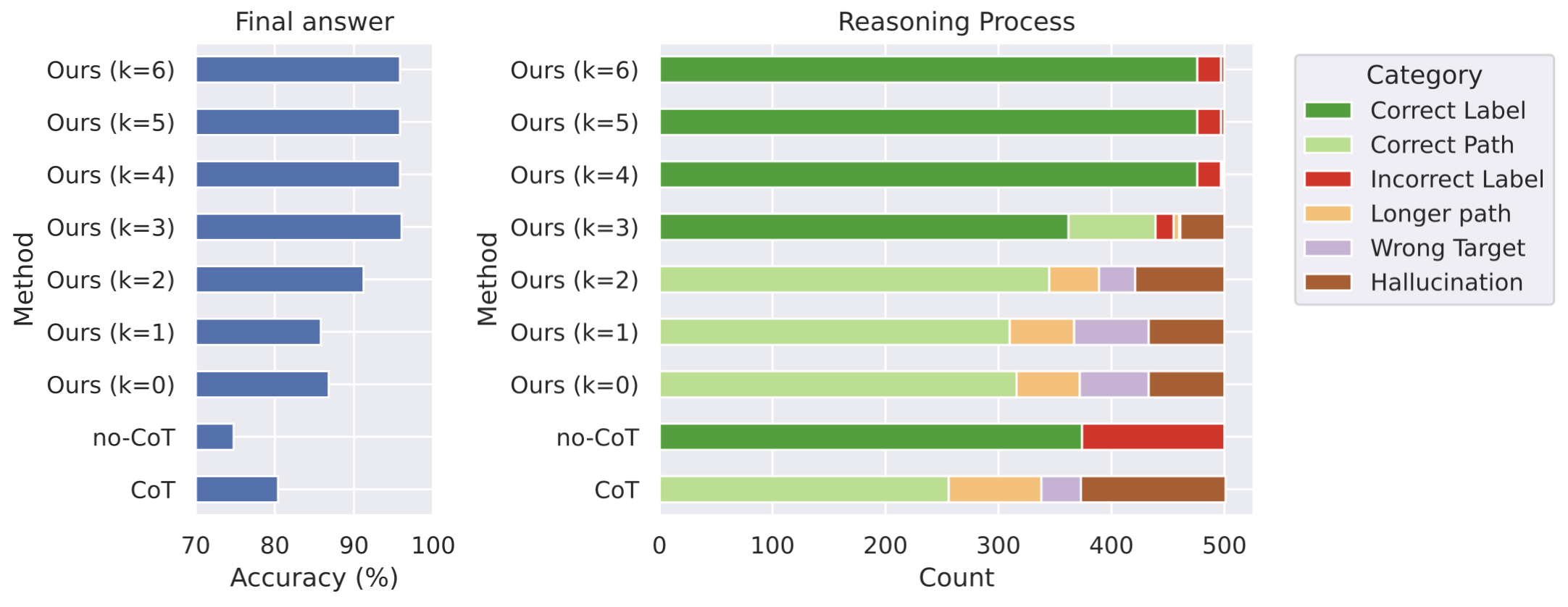}
    \caption{The accuracy of final answer (left) and reasoning process (right) of multiple variants of \ours and baselines on ProsQA.}
    \label{fig:analysis}
\end{figure}

Figure~\ref{fig:analysis} presents a comparative analysis of various reasoning methods evaluated on ProsQA. The model trained using \textit{CoT} frequently hallucinates non-existent edges or outputs paths leading to incorrect targets, resulting in lower answer accuracy. In contrast, \ours, which leverages continuous space reasoning, demonstrates improved accuracy as it utilizes an increasing number of continuous thoughts. Additionally, the rate of correct reasoning processes (indicated by ``Correct Label'' and ``Correct Path'') significantly increases. At the same time, there is a notable reduction in instances of ``Hallucination'' and ``Wrong Target,'' issues that typically emerge when the model makes mistakes early in the reasoning process.

An intuitive demonstration of the limitations of reasoning in language space is provided by the case study depicted in Figure~\ref{fig:interpret}. As shown, models operating in language space often fail to plan ahead or backtrack. Once they commit to an incorrect path, they either hallucinate unsupported edges or terminate with irrelevant conclusions. In contrast, latent reasoning avoids such premature commitments by enabling the model to iteratively refine its decisions across multiple reasoning steps. This flexibility allows the model to progressively eliminate incorrect options and converge on the correct answer, ultimately resulting in higher accuracy.


\subsection{Interpreting the Latent Reasoning as Tree Search}
\label{sec:interpret}

\begin{figure}[H]
    \centering
    \vspace{-15pt}
    \includegraphics[width=1\linewidth]{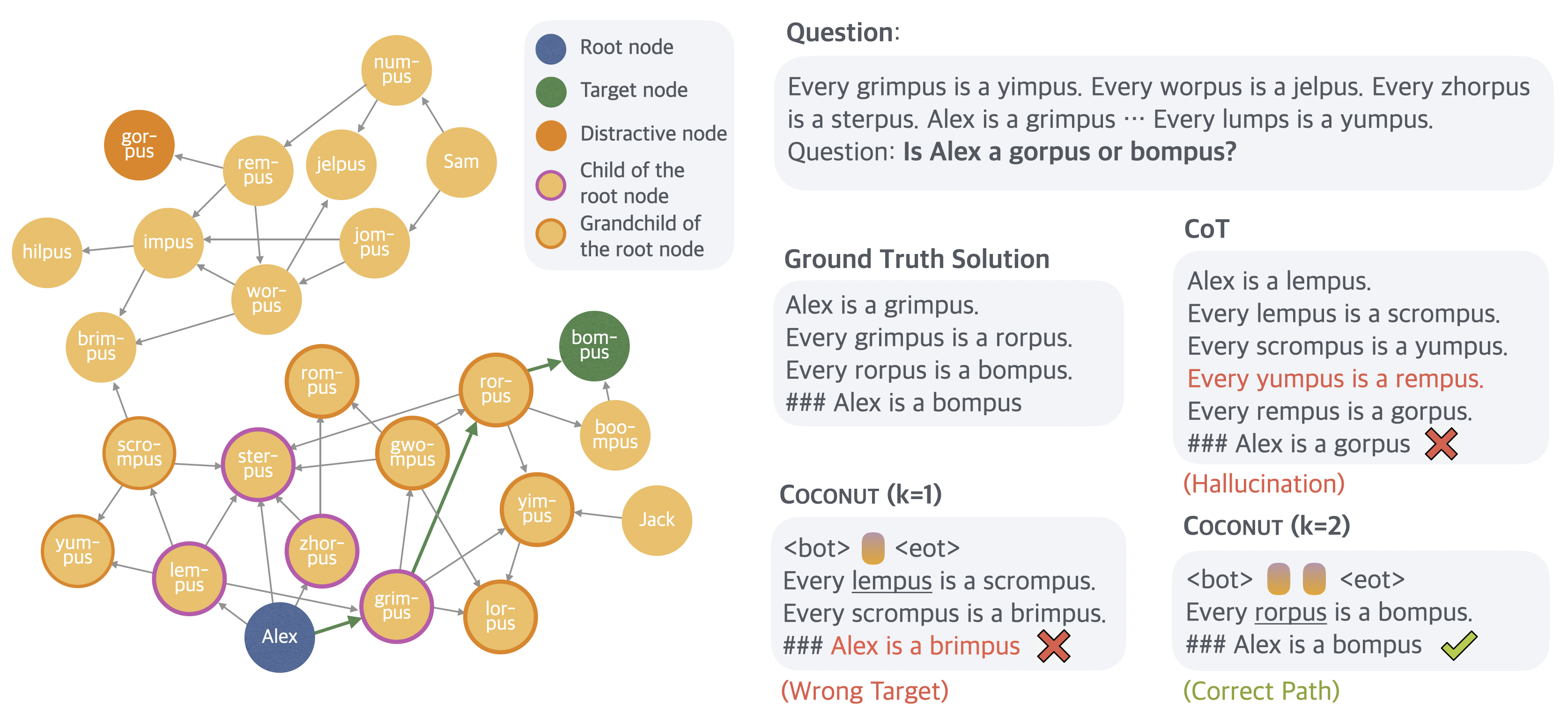}
    
    \caption{A case study of ProsQA. The model trained with \textit{CoT} hallucinates an edge (\textit{Every yumpus is a rempus}) after getting stuck in a dead end. \ours (k=1) outputs a path that ends with an irrelevant node. \ours (k=2) solves the problem correctly.
    }
    \label{fig:interpret}
    \centering
    
    \vspace{15pt}
    \includegraphics[width=\linewidth]{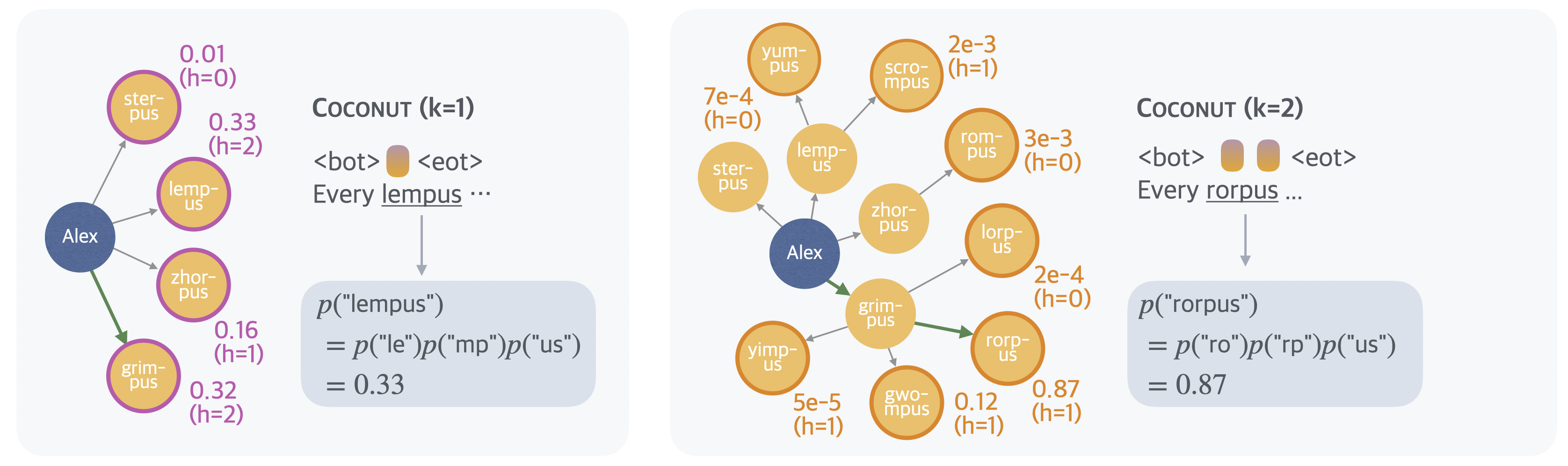}
    
    \caption{An illustration of the latent search trees. The example is the same test case as in Figure~\ref{fig:interpret}. The height of a node (denoted as $h$ in the figure) is defined as the longest distance to any leaf nodes in the graph. We show the probability of the first concept predicted by the model following latent thoughts (e.g., ``lempus'' in the left figure). It is calculated as the multiplication of the probability of all tokens within the concept conditioned on previous context (omitted in the figure for brevity). This metric can be interpreted as an implicit value function estimated by the model, assessing the potential of each node leading to the correct answer.}
    \label{fig:height}
    \vspace{-10pt}
\end{figure}

To better understand \ours, we probe the latent reasoning process by forcing the model to explicitly generate language reasoning steps following intermediate continuous thoughts (Figure~\ref{fig:height}). Using the example presented in Figure~\ref{fig:interpret}, at the initial reasoning step, the model must select which immediate child node of ``Alex'' to consider next, specifically from the set \{``lempus'', ``sterpus'', ``zhorpus'', ``grimpus''\}. The distribution over these candidate next steps is visualized in Figure~\ref{fig:height}, left. In the subsequent reasoning step, these nodes expand further into an extended set of potential paths, including all grandchildren of ``Alex'' (Figure~\ref{fig:height}, right).

We define the predicted probability of a concept following continuous thoughts as a value function (Figure~\ref{fig:height}), estimating each node's potential for reaching the correct target. Interestingly, the reasoning strategy employed by \ours is not greedy search: while ``lempus'' initially has the highest value (0.33) at the first reasoning step (Figure~\ref{fig:height}, left), the model subsequently assigns the highest value (0.87) to ``rorpus,'' a child of ``grimpus,'' rather than following ``lempus'' (Figure~\ref{fig:height}, right). This characteristic resembles a breadth-first search (BFS) approach, contrasting sharply with the greedy decoding typical of traditional CoT methods. The inherent capability of continuous representations to encode multiple candidate paths enables the model to avoid making immediate deterministic decisions.  Importantly, this tree search pattern is not limited to the illustrated example, but constitutes a fundamental mechanism underlying the consistent improvement observed with larger values of $k$ in \ours.

\begin{figure*}
    \centering
    \includegraphics[width=0.9\linewidth]{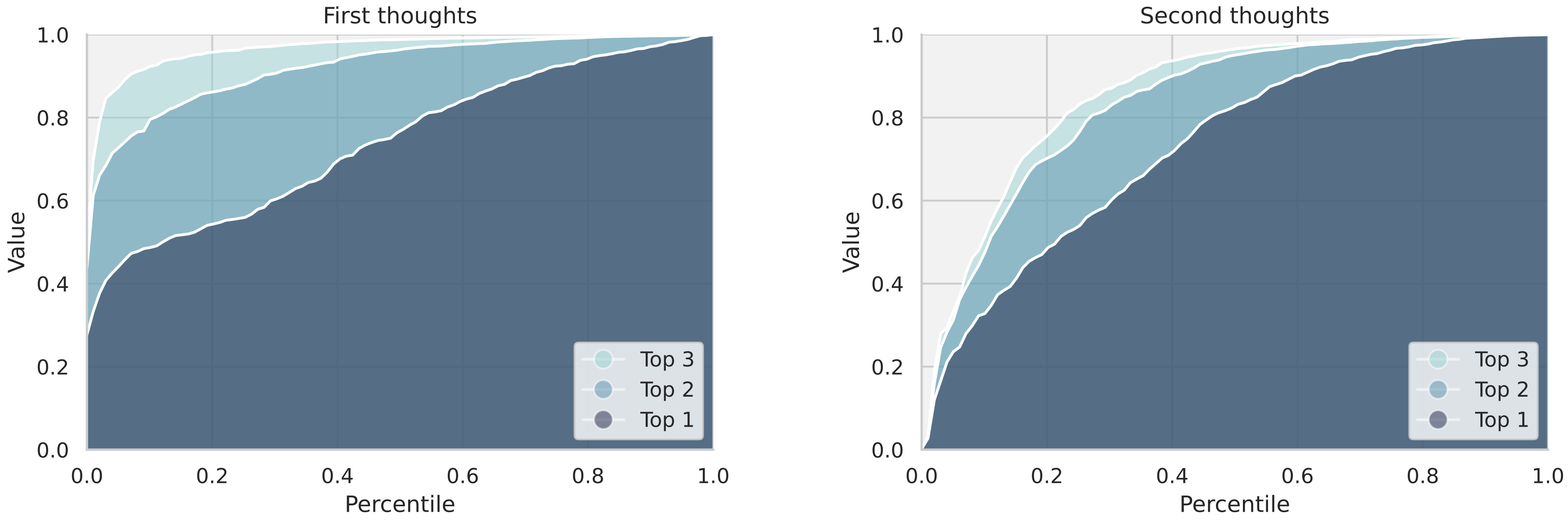}
    \vspace{-5pt}
    \caption{Analysis of parallelism in the first two steps of the latent tree search. The three curves in each panel depict the cumulative value of the top-1, top-2, and top-3 candidate nodes.}
    \label{fig:percentile}
\end{figure*}

Figure~\ref{fig:percentile} presents an analysis of the parallelism in the model's latent reasoning across the first and second thoughts. For the first thoughts (left panel), the cumulative values of the top-1, top-2, and top-3 candidate nodes are computed and plotted against their respective percentiles across the test set. The noticeable gaps between the three lines indicate that the model maintains significant diversity in its reasoning paths at this stage, suggesting a broad exploration of alternative possibilities. In contrast, the second thoughts (right panel) show a narrowing of these gaps. This trend suggests that the model transitions from parallel exploration to more focused reasoning in the second latent reasoning step, likely as it gains more certainty about the most promising paths. 

\subsection{Why is Latent Space Better for Planning?}
\label{sec:height}

Building upon the tree search perspective, we further examine why latent reasoning benefits planning tasks—specifically, why maintaining multiple candidate paths and postponing deterministic decisions enhances reasoning performance. Our hypothesis is that nodes explored in the early reasoning stages are inherently more challenging to evaluate accurately because they are farther from the final target nodes. In contrast, nodes positioned closer to potential targets, having fewer subsequent exploration possibilities, can be assessed accurately with higher confidence.

\begin{wrapfigure}{r}{0.40\linewidth}
    \vspace{-3em}
    \includegraphics[width=\linewidth]{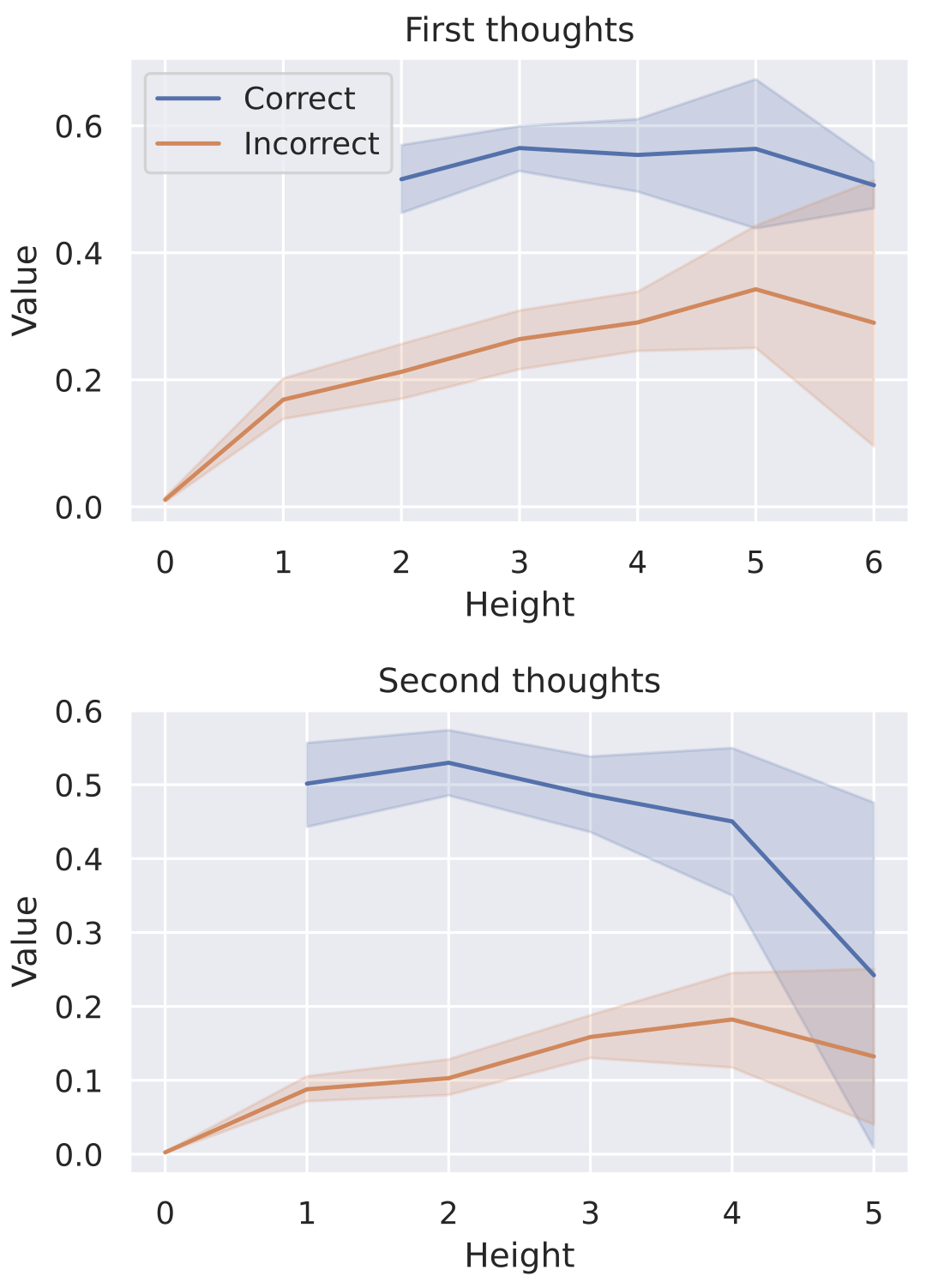}
    \caption{The correlation between the predicted value of correct/incorrect nodes and their heights.}
    \vspace{-5em}
    \label{fig:height_pred}
\end{wrapfigure}

To systematically test this, we define the height of a node as its shortest distance to any leaf node and analyze the relationship between node height and the model's estimated value. Ideally, a correct node—one that can lead to the target node—should receive a high estimated value, whereas an incorrect node—one that cannot lead to the target node—should receive a low value. Empirical results across the test set (Figure~\ref{fig:height_pred}) support our hypothesis: nodes with lower heights consistently receive more accurate and definitive probability evaluations. Conversely, nodes with greater heights exhibit more ambiguous evaluations, reflecting increased uncertainty.

These findings underscore the advantage of latent space reasoning. By delaying deterministic decisions and allowing exploration to proceed toward terminal states, latent reasoning significantly enhances the model’s ability to differentiate correct paths from incorrect ones, thereby improving performance on complex, planning-intensive tasks compared to traditional greedy methods. 

\section{Empirical Results with \ours}
\label{sec:experiment}
After analyzing the promising parallel search pattern of \ours, we validate the feasibility of LLM reasoning in a continuous latent space through more comprehensive experiments, highlighting its better reasoning efficiency over language space, as well as its potential to enhance the model's expressivity with test-time scaling.

\begin{table*}
    \centering
    \label{tab:main}
    \begin{tabular}{@{}rcccccc}
    
    \toprule
    \multirow{2}{*}[-2pt]{Method} & \multicolumn{2}{c}{GSM8k} & \multicolumn{2}{c}{ProntoQA} & \multicolumn{2}{c}{\dataset}\\
    \cmidrule{2-7}
          & Acc. (\%) & \# Tokens & Acc. (\%) & \# Tokens & Acc. (\%) & \# Tokens \\
        \midrule
        CoT  & 42.9{\scriptsize$\ \pm$0.2} & 25.0 &  98.8{\scriptsize$\ \pm$0.8} & 92.5 &  77.5{\scriptsize$\ \pm$1.9} & 49.4 \\
        \midrule
        No-CoT  & 16.5{\scriptsize$\ \pm$0.5} & 2.2 & 93.8{\scriptsize$\ \pm$0.7}  & 3.0 & 76.7{\scriptsize$\ \pm$1.0} & 8.2 \\
        iCoT  & 30.0$^*$ &  2.2 &  99.8{\scriptsize$\ \pm$0.3} &  3.0 & 98.2{\scriptsize$\ \pm$0.3} & 8.2 \\
        Pause Token &  16.4{\scriptsize$\ \pm$1.8} & 2.2 & 77.7{\scriptsize$\ \pm$21.0} & 3.0 &  75.9{\scriptsize$\ \pm$0.7} & 8.2 \\
        \midrule
        
        \ours (Ours) & 34.1{\scriptsize$\ \pm$1.5} & 8.2  & 99.8{\scriptsize$\ \pm$0.2} & 9.0 &  97.0{\scriptsize$\ \pm$0.3} & 14.2 \\
        
        - \textit{w/o curriculum} & 14.4{\scriptsize$\ \pm$0.8} & 8.2 & 52.4{\scriptsize$\ \pm$0.4} & 9.0 &  76.1{\scriptsize$\ \pm$0.2} & 14.2 \\
        - \textit{w/o thought} & 21.6{\scriptsize$\ \pm$0.5}  & 2.3 &  99.9{\scriptsize$\ \pm$0.1} & 3.0 &  95.5{\scriptsize$\ \pm$1.1} & 8.2 \\
        - \textit {pause as thought} & 24.1{\scriptsize$\ \pm$0.7} & 2.2 & 100.0{\scriptsize$\ \pm$0.1} & 3.0 &  96.6{\scriptsize$\ \pm$0.8} & 8.2 \\
    \bottomrule 
    \end{tabular}
     \small
    \caption{Results on three datasets: GSM8k, ProntoQA and ProsQA. Higher accuracy indicates stronger reasoning ability, while generating fewer tokens indicates better efficiency. $^*$The result is from \citet{deng2024explicit}.}
    \label{tab:main}
\end{table*}

\subsection{Experimental Setup}

\noindent\textbf{Math Reasoning.}
We use GSM8k~\citep{cobbe2021training} as the dataset for math reasoning. It consists of grade school-level math problems. To train the model, we use a synthetic dataset generated by~\citet{deng2023implicit}. We use two continuous thoughts for each reasoning step (i.e., $c=2$). The model goes through 3 stages besides the initial stage. We then include an additional stage where still $3\times c$ continuous thoughts are used as in the previous stage, but with all the remaining language reasoning chain removed. This handles the long-tail distribution of reasoning chains longer than 3 steps. We train the model for 6 epochs in the initial stage, and 3 epochs in each remaining stage. 

\noindent\textbf{Logical Reasoning.} Logical reasoning involves the proper application of known conditions to prove or disprove a conclusion using logical rules. We use the ProntoQA~\citep{saparov2022language} dataset, and our newly proposed ProsQA dataset, which is more challenging due to more distracting branches. We use one continuous thought for every reasoning step (i.e., $c=1$). The model goes through 6 training stages in addition to the initial stage, because the maximum number of reasoning steps is 6 in these two datasets. The model then fully reasons with continuous thoughts to solve the problems in the last stage. We train the model for 5 epochs per stage.

For all datasets, after the standard schedule, the model stays in the final training stage, until reaching 50 epochs. We select the checkpoint based on the accuracy on the validation set. For inference, we manually set the number of continuous thoughts to be consistent with their final training stage. We use greedy decoding for all experiments.

\subsection{Baselines and Variants of \ours}
We consider the following baselines: (1) \textit{CoT}, and (2) \textit{No-CoT}, which were introduced in Section~\ref{sec:understanding}. (3) \textit{iCoT}~\citep{deng2024explicit}: The model is trained with language reasoning chains and follows a carefully designed schedule that ``internalizes'' CoT. As the training goes on, tokens at the beginning of the reasoning chain are gradually removed until only the answer remains. During inference, the model directly predicts the answer. (4) \textit{Pause token}~\citep{goyal2023think}: The model is trained using only the question and answer without a reasoning chain. However, different from \textit{No-CoT}, special \texttt{<pause>} tokens are inserted between the question and answer, which provides the model with additional computational capacity to derive the answer. The number of \texttt{<pause>} tokens is set the same as continuous thoughts in \ours.

We also evaluate some variants of \ours: (1) \textit{w/o curriculum}, which directly trains the model in the last stage. The model uses continuous thoughts to solve the whole problem. (2) \textit{w/o thought}: We keep the multi-stage training, but don't add any continuous latent thoughts. While this is similar to \textit{iCoT} in the high-level idea, the exact training schedule is set to be consistent with \ours, instead of \textit{iCoT}, for a strict comparison. (3) \textit{Pause as thought}: We use special \texttt{<pause>} tokens to replace the continuous thoughts, and apply the same multi-stage training curriculum as \ours.

\subsection{Results and Discussion}

\label{sec:result}

We show the overall results in Table~\ref{tab:main}. Using continuous thoughts effectively enhances LLM reasoning over the No-CoT baseline. For example, by using 6 continuous thoughts, \ours achieves 34.1\% accuracy on GSM8k, which significantly outperforms \textit{No-CoT} (16.5\%). We highlight several key findings below.

\begin{wrapfigure}{r}{0.36\linewidth}

    \centering
    \includegraphics[width=\linewidth]{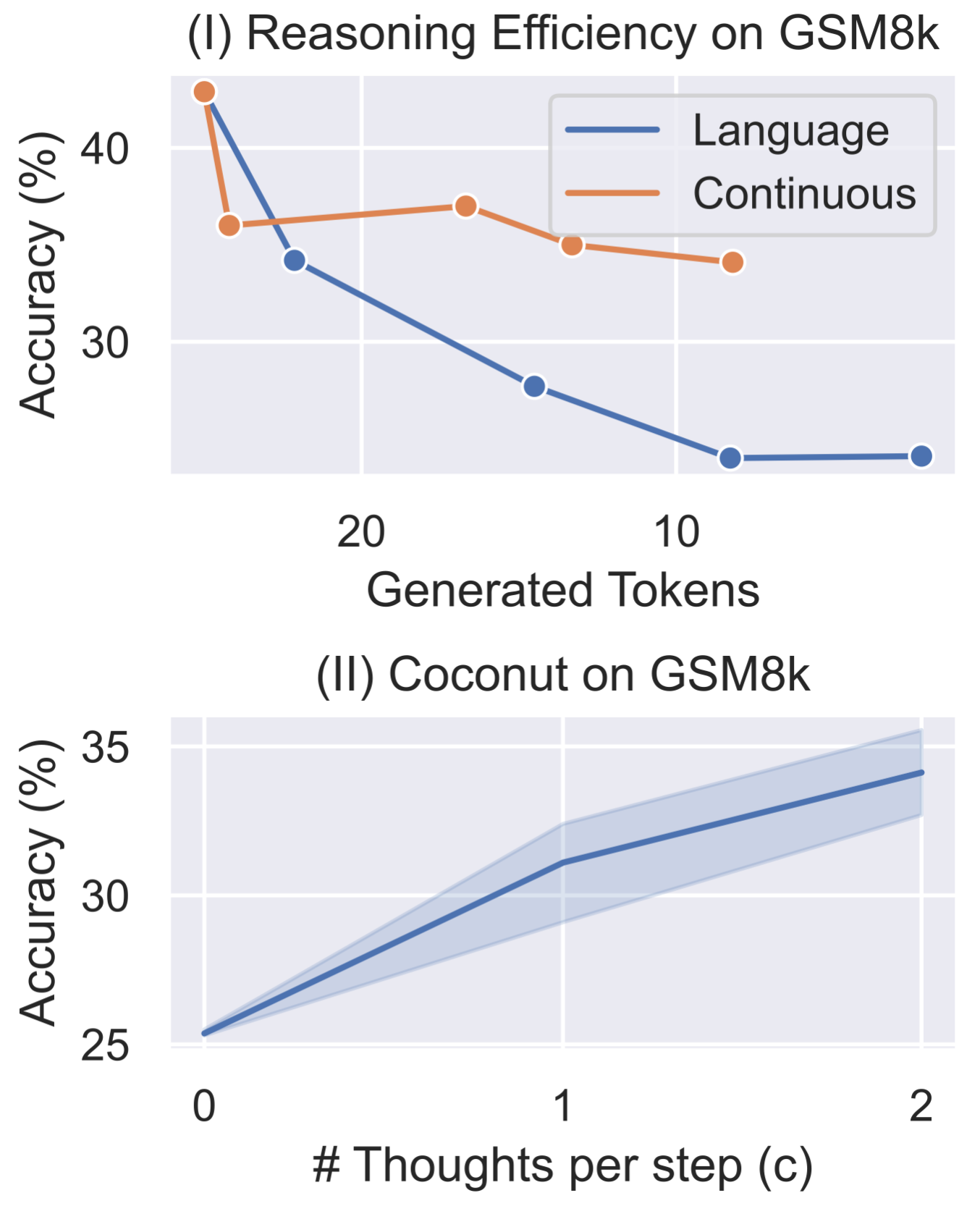}
    \vspace{-1em}
    \caption{Efficiency comparison of reasoning space and \ours with different $c$.}
    \vspace{-2em}
    \label{fig:merge}
\end{wrapfigure}

\noindent \textbf{``Chaining'' continuous thoughts enhances reasoning.} Language CoT proves to increase the effective depth of LLMs and enhance their expressiveness~\citep{feng2023towards}. Thus, generating more tokens serves as a way to inference-time scaling for reasoning~\citep{guo2025deepseek, snell2024scaling}. This desirable property holds naturally for \ours too. On GSM8k, \ours outperformed other architectures trained with similar strategies, including \ours (\textit{pause as thought}) and \ours (\textit{w/o thought}). Particularly, it surpasses the latest baseline \textit{iCoT}~\citep{deng2024explicit}, which requires a more carefully designed training schedule.

Additionally, we experimented with adjusting the hyperparameter $c$, which controls the number of latent thoughts corresponding to one language reasoning step (Figure~\ref{fig:merge}, II). As we increased $c$ from 0 to 1 to 2, the model's performance steadily improved.\footnote{We discuss the case of larger $c$ in Appendix~\ref{sec:larger_c}.} This further validates the potential of continuous thoughts to scale up to harder problems. In two other synthetic tasks, we found that the variants of \ours (\textit{w/o thoughts} or \textit{pause as thought}), and the \textit{iCoT} baseline also achieve impressive accuracy. This indicates that the model's computational capacity may not be the bottleneck in these tasks. In contrast, GSM8k involves more complex contextual understanding and modeling, placing higher demands on computational capability.

\begin{wrapfigure}{r}{0.46\linewidth}
    \centering
    \includegraphics[width=\linewidth]{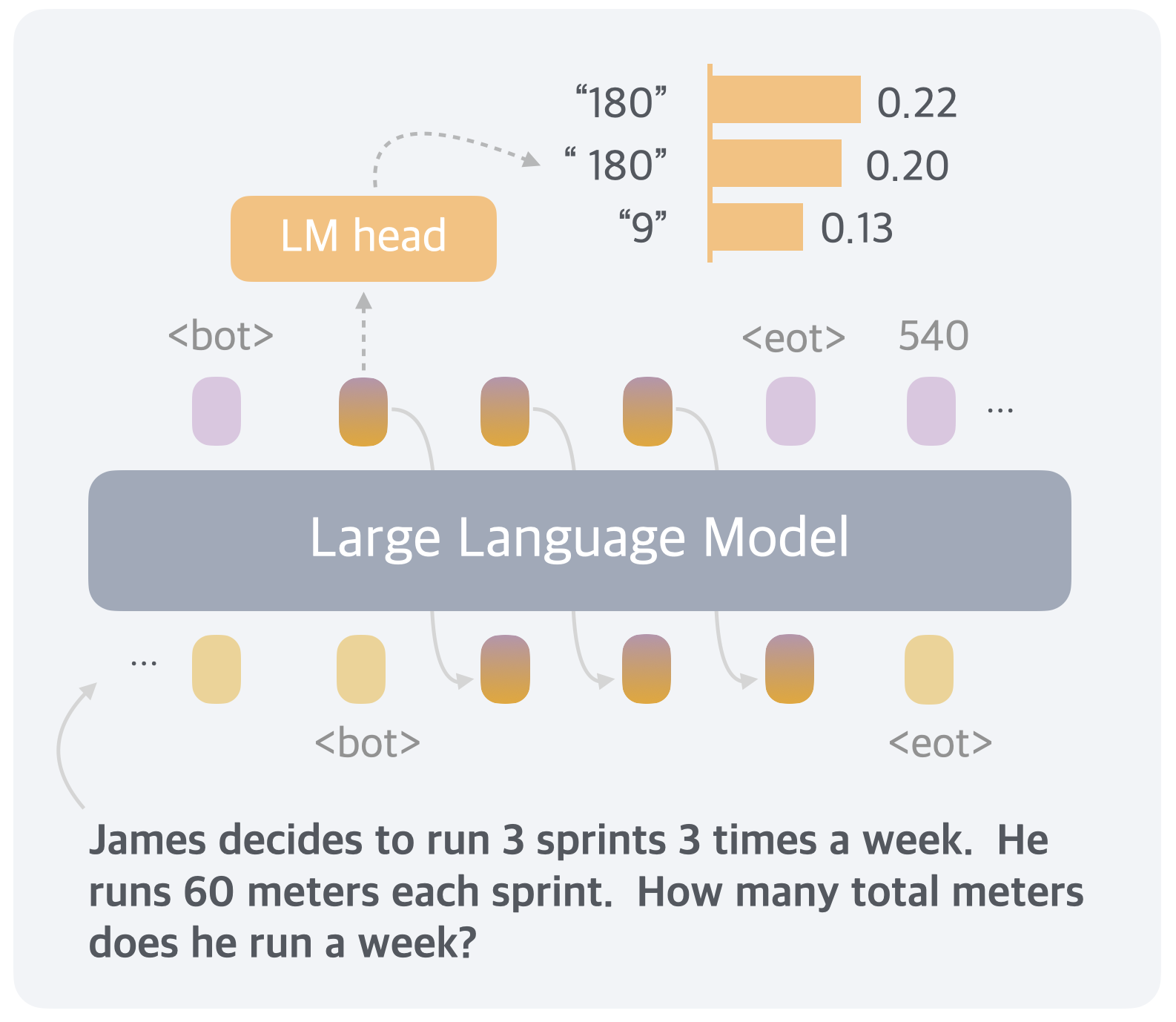}
    \caption{Decoding a continuous thought into language tokens in a math word problem. The decoded tokens correspond to intermediate variables that help solve the problem.}
    \vspace{-2em}
    \label{fig:case}
\end{wrapfigure}

\noindent \textbf{Continuous thoughts are efficient representations of reasoning.} 
Compared to traditional CoT, \ours generates fewer tokens while achieving higher accuracy on ProntoQA and ProsQA (Table~\ref{tab:main}). Although \ours does not surpass \textit{CoT} on GSM8k, it offers a superior trade-off between reasoning efficiency and accuracy (Figure~\ref{fig:merge}, I). To illustrate this, we train a series of CoT models that progressively ``internalize''~\citep{deng2024explicit} the initial $m=\{0, 1, 2, 3, \text{ALL}\}$ reasoning steps, and plot their accuracy versus the number of generated tokens (labeled as ``language'' in the figure). These models quickly lose accuracy as more reasoning steps are skipped. In contrast, by applying \ours training strategy—replacing each language reasoning step with two continuous thoughts—the accuracy drop is substantially mitigated, maintaining higher performance even when fewer tokens are generated. Another interesting observation is that, when we decode the first continuous thought, it often corresponds to possible intermediate variables in the calculation (Figure~\ref{fig:case}). This also suggests that the continuous thoughts are more efficient representations of reasoning.

\noindent \textbf{The LLM still needs guidance to learn latent reasoning.} 
In the ideal case, the model should learn the most effective continuous thoughts automatically through gradient descent on questions and answers (i.e., \ours \textit{w/o curriculum}). However, from the experimental results, we found  the models trained this way do not perform any better than no-CoT.

On the contrary, with the multi-stage curriculum, \ours is able to achieve top performance across various tasks. The multi-stage training also integrates well with pause tokens (\ours - \textit{pause as thought}). Despite using the same architecture and similar multi-stage training objectives, we observed a small gap between the performance of \textit{iCoT} and \ours (\textit{w/o thoughts}). The finer-grained removal schedule (token by token) and a few other tricks in \textit{iCoT} may ease the training process. We leave combining \textit{iCoT} and \ours as future work. While the multi-stage training used for \ours has proven effective, further research is definitely needed to develop better and more general strategies for learning reasoning in latent space, especially without the supervision from language reasoning chains.

\section{Conclusion}

In this paper, we introduce \ours, a new paradigm for reasoning in continuous latent space. Experiments demonstrate that \ours effectively enhances LLM performance across a variety of reasoning tasks. Reasoning in latent space gives rise to advanced emergent behaviors, where continuous thoughts can represent multiple alternative next steps. This enables the model to perform BFS over possible reasoning paths, rather than prematurely committing to a single deterministic trajectory as in language space CoT reasoning. Further research is needed to refine and scale latent reasoning to pretraining, which could improve generalization across a broader range of reasoning challenges. We hope our findings will spark continued exploration into latent reasoning, ultimately advancing the development of more capable machine reasoning systems.

\newpage
{\Large \sffamily Acknowledgement\par}
The authors express their sincere gratitude to Jihoon Tack for his valuable discussions throughout the course of this work.

\bibliographystyle{assets/plainnat}
\bibliography{paper}

\clearpage
\newpage
\beginappendix
\section{Datasets}
\label{sec:dataset-appendix}

\subsection{Examples}
We provide some examples of the questions and CoT solutions for the datasets used in our experiments.

\begin{tcolorbox}[title=GSM8k, colback=white]
\texttt{Question = "John cuts his grass to 2 inches.  It grows .5 inches per month.  When it gets to 4 inches he cuts it back down to 2 inches.  It cost \$100 to get his grass cut.  How much does he pay per year?" \\ Steps = ["<<4-2=2>>", "<<2/.5=4>>", "<<12/4=3>>", "<<100*3=300>>"]\\
Answer = "300"
}
\end{tcolorbox}

\begin{tcolorbox}[title=ProntoQA, colback=white]
\texttt{Question = "Brimpuses are not luminous. Shumpuses are amenable. Each yumpus is a lorpus. Gorpuses are shumpuses. Each zumpus is a grimpus. Gorpuses are rompuses. Dumpuses are not floral. Lempuses are cold. Brimpuses are impuses. Every lorpus is floral. Every rompus is transparent. Grimpuses are muffled. Rompuses are yumpuses. Rompuses are wumpuses. Zumpuses are fast. Wumpuses are bitter. Every sterpus is orange. Each lorpus is a vumpus. Yumpuses are feisty. Each yumpus is a lempus. Gorpuses are snowy. Zumpuses are gorpuses. Every lorpus is a sterpus. Stella is a brimpus. Stella is a zumpus. True or false: Stella is not floral." \\ Steps = ["Stella is a zumpus. Zumpuses are gorpuses.", "Stella is a gorpus. Gorpuses are rompuses.", "Stella is a rompus. Rompuses are yumpuses.", "Stella is a yumpus. Each yumpus is a lorpus.", "Stella is a lorpus. Every lorpus is floral.", "Stella is floral."]\\
Answer = "False"
}
\end{tcolorbox}

\begin{tcolorbox}[title=ProsQA, colback=white]
\texttt{Question = "Every shumpus is a rempus. Every shumpus is a yimpus. Every terpus is a fompus. Every terpus is a gerpus. Every gerpus is a brimpus. Alex is a rempus. Every rorpus is a scrompus. Every rorpus is a yimpus. Every terpus is a brimpus. Every brimpus is a lempus. Tom is a terpus. Every shumpus is a timpus. Every yimpus is a boompus. Davis is a shumpus. Every gerpus is a lorpus. Davis is a fompus. Every shumpus is a boompus. Every shumpus is a rorpus. Every terpus is a lorpus. Every boompus is a timpus. Every fompus is a yerpus. Tom is a dumpus. Every rempus is a rorpus. Is Tom a lempus or scrompus?" \\ Steps = ["Tom is a terpus.", "Every terpus is a brimpus.", "Every brimpus is a lempus."]\\
Answer = "Tom is a lempus."
}
\end{tcolorbox}

\subsection{Construction of ProsQA}

To construct the dataset, we first compile a set of typical entity names, such as “Alex” and “Jack,” along with fictional concept names like “lorpus” and “rorpus,” following the setting of ProntoQA~\citep{saparov2022language}. Each problem is structured as a binary question: “Is [Entity] a [Concept A] or [Concept B]?” Assuming [Concept A] is the correct answer, we build a directed acyclic graph (DAG) where each node represents an entity or a concept. The graph is constructed such that a path exists from [Entity] to [Concept A] but not to [Concept B].

Algorithm~\ref{alg:prosqa} describes the graph construction process. The DAG is incrementally built by adding nodes and randomly connecting them with edges. To preserve the validity of the binary choice, with some probability, we enforce that the new node cannot simultaneously serve as a descendant to both node $0$ and $1$. This separation maintains distinct families of nodes and balances their sizes to prevent model shortcuts.

After the graph is constructed, nodes without parents are assigned entity names, while other nodes receive concept names. To formulate a question of the form “Is [Entity] a [Concept A] or [Concept B]?”, we designate node $0$ in the graph as [Entity], a leaf node labeled $1$ as [Concept A], and a leaf node labeled $2$ as [Concept B]. This setup ensures a path from [Entity] to [Concept A] without any connection to [Concept B], introducing a moderately complex reasoning path. Finally, to avoid positional biases, [Concept A] and [Concept B] are randomly permuted in each question.

\begin{algorithm}
\caption{Graph Construction for ProsQA}\label{alg:prosqa}
\begin{algorithmic}

\State  $edges\gets \{\}$
\State  $nodes\gets \{0, 1\}$
\State  $labels\gets \{0: 1, 1:2\}$ \\ \Comment{Labels: 1 (descendant of node 0), 2 (descendant of node 1), 3 (both), 0 (neither).}
\State  $groups\gets \{0: \{\}, 1: \{0\}, 2:\{1\}, 3:\{\}\}$

\State $idx\gets 2$
\While{$idx < N$}

    \Comment{For each new node, randomly add edges from existing nodes}
    
    \State $n\_in\_nodes \gets \text{poisson}(1.5)$
    \State $rand \gets \text{random}()$

    \If{$rand \le 0.35$}
        \State $candidates \gets groups[0] \cup groups[1]$
        \Comment{Cannot be a descendant of node 1.}
    \ElsIf{$rand \le 0.7$}
        \State $candidates \gets groups[0] \cup groups[2]$
        \Comment{Cannot be a descendant of node 0.}
    \Else
        \State $candidates \gets nodes$
    \EndIf
    \State $n\_in\_nodes \gets \min(\text{len}(candidates), n\_in\_nodes)$
    \State $weights \gets [\text{depth\_to\_root}(c) \cdot 1.5 + 1 \;\forall c \in candidates]$ \\
    \Comment{Define sampling weights to prioritize deeper nodes.} \\
    \Comment{This way, the solution reasoning chain is expected to be longer.}

    \State $in\_nodes \gets \text{random\_choice}(candidates, n\_in\_nodes, \text{prob} = weights / \text{sum}(weights))$
    \State $cur\_label \gets 0$
    
    \For{$in\_idx \in in\_nodes$}
        \State $cur\_label \gets cur\_label \mid labels[in\_idx]$
        \Comment{Update label using bitwise OR.}
        \State $edges.\text{append}((in\_idx, idx))$
    \EndFor
    
    \State $groups[cur\_label].\text{append}(idx)$
    \State $labels[idx] \gets cur\_label$
    \State $nodes \gets nodes \cup \{idx\}$
    \State $idx \gets idx + 1$

\EndWhile
\end{algorithmic}
\end{algorithm}

\begin{table}[]

    \centering
    \begin{tabular}{c|c|c|c}
    \toprule
         \# Nodes & \# Edges & Len. of Shortest Path & \# Shortest Paths \\
         \midrule
           23.0 & 36.0 & 3.8 & 1.6 \\
          \bottomrule
    \end{tabular}
    \captionsetup{justification=centering}
    \caption{Statistics of the graph structure in ProsQA.}
    \label{tab:prosqa}
\end{table}

\subsection{Statistics}

We show the size of all datasets in Table~\ref{tab:stats}.

\begin{table}[]

    \centering
    \begin{tabular}{r|c|c|c}
    \toprule
         Dataset & Training & Validation & Test \\
         \midrule
           GSM8k & 385,620 & 500 & 1319 \\
           ProntoQA & 9,000 & 200 & 800 \\
           ProsQA & 17,886 & 300 & 500 \\
          \bottomrule
    \end{tabular}
    \captionsetup{justification=centering}
    \caption{Statistics of the datasets.}
    \label{tab:stats}
\end{table}

\section{Clock-Time Reasoning Efficiency Metric} \label{sec:clock}

We present a clock-time comparison to evaluate reasoning efficiency. The reported values represent the average inference time per test case (in seconds), with a batch size of 1, measured on an Nvidia A100 GPU. For the no-CoT and CoT baselines, we employ the standard generate method from the \texttt{transformers}\footnote{\url{https://github.com/huggingface/transformers}} library. Our results show that clock time is generally proportional to the number of newly generated tokens, as detailed in Table~\ref{tab:main}.

\begin{table}[h!]
    \centering
    \label{tab:efficiency}
    \begin{tabular}{lccc}
        \toprule
        Method & GSM8k & ProntoQA & ProsQA \\
        \midrule
        No-CoT & 0.03 & 0.03 & 0.08 \\
        CoT    & 0.26 & 0.85 & 0.47 \\
        \ours   & 0.09 & 0.11 & 0.15 \\
        \bottomrule
        
    \end{tabular}
    
    \captionsetup{justification=centering}
    \caption{Inference time (in seconds) comparison across tasks and methods.}
\end{table}

\section{More Discussion}
\subsection{Using More Continuous Thoughts}

\label{sec:larger_c}
In Figure~\ref{fig:merge} (II), we present the performance of \ours on GSM8k using $c \in \{0, 1, 2\}$. When experimenting with $c=3$, we observe a slight performance drop accompanied by increased variance. Analysis of the training logs indicates that adding three continuous thoughts at once -- particularly during the final stage transition -- leads to a sharp spike in training loss, causing instability. Future work will explore finer-grained schedules, such as incrementally adding continuous thoughts one at a time while removing fewer language tokens, as in iCoT~\citep{deng2024explicit}. Additionally, combining language and latent reasoning—e.g., generating the reasoning skeleton in language and completing the reasoning process in latent space—could provide a promising direction for improving performance and stability.

\subsection{\ours with Larger Models}

We experimented with \ours on GSM8k using Llama 3.2-3B and Llama 3-8B~\citep{dubey2024llama} with $c=1$. We train them for 3 epochs in Stage 0, followed by 1 epoch per subsequent stage. The results are shown in Table~\ref{tab:coconut-large-models}.

\begin{table}[h]
\centering
\begin{tabular}{lcc}
\toprule
\textbf{Model} & \textbf{no-CoT} & \textbf{\ours (Ours)} \\ 
\midrule
Llama 3.2-3B & 26.0 & 31.7 \\
Llama 3-8B   & 42.2 & 43.6 \\
\bottomrule
\end{tabular}
\caption{Experimental results of applying \ours to larger Llama models. We report performance comparisons between models without CoT reasoning (no-CoT) and our proposed \ours method.}
\label{tab:coconut-large-models}
\end{table}

We observe consistent performance gains across both Llama 3.2-3B and Llama 3-8B models compared to the no-CoT baseline, though these improvements are not as pronounced as those previously demonstrated with GPT-2. One possible reason is that larger models have already undergone extensive language-focused pre-training, making the transition to latent reasoning more challenging.

We emphasize that the primary goal of this paper is to highlight the promising attributes of latent-space reasoning and to initiate exploration in this new direction. Universally surpassing language-based CoT likely requires significant research efforts dedicated to \textbf{latent space pre-training}. We are encouraged by recent progress in this area~\citep{geiping2025scaling, barrault2024large, gladstone2025energy}. While these recent models provide scalable methods for latent representation learning, the latent spaces have not yet been explicitly optimized for reasoning. Integrating these recent advancements with \ours presents an exciting and promising avenue for future research.

\end{document}